# Reconstructing Pompeian Households

**David Mimno**
Department of Computer Science
Princeton University
Princeton, NJ 08540
mimno@cs.umass.edu

## Abstract

A database of objects discovered in houses in the Roman city of Pompeii provides a unique view of ordinary life in an ancient city. Experts have used this collection to study the structure of Roman households, exploring the distribution and variability of tasks in architectural spaces, but such approaches are necessarily affected by modern cultural assumptions. In this study we present a data-driven approach to household archeology, treating it as an unsupervised labeling problem. This approach scales to large data sets and provides a more objective complement to human interpretation.

## 1 Introduction

Over the past century the goal of archeology has shifted from finding objects of artistic value to reconstructing the details of day-to-day life in ancient cultures. Reconstructing daily life is difficult as the evidence that would be useful was generally considered commonplace and unworthy of preservation by people at the time. Objects are gradually moved or disposed of; structures are renovated, repurposed or recycled for building materials. These ongoing processes make it difficult to make statements about specific points in time based on material evidence. A remarkable exception to this pattern is the Roman city of Pompeii in southern Italy, which was was covered in volcanic ash during an eruption of Mt. Vesuvius in 79 AD. As a result of its sudden, violent destruction, many aspects of day-to-day life in first-century Roman Italy were preserved in place as they existed on a particular day.

The study of Pompeian households has until recently been dominated by analyses of architectural patterns and wall paintings, as these are easily available for study. Only in the past few years has a database of objects, which are removed for conservation immediately upon excavation, been compiled and made available by Allison [2]. This database, which is available on-line,[1] contains more than 6000 artifact records for finds in 30 architecturally similar "*atrium*-style" houses in Pompeii. For each artifact, the database specifies a type from 240 typological categories (coin, amphora, etc.) and a find location from 574 rooms. Allison has used data about artifacts in their original context to challenge many common assumptions about the function of particular types of object, the use of particular spaces, and the consistency of patterns of use across different houses.

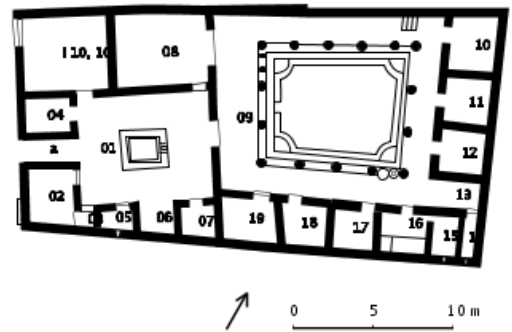

Figure 1: **A Roman house**, the Casa degli Amanti from Pompeii. The entrance is at left, opening onto the *atrium* (01), which leads to a colonnaded garden (09). Image from http://www.stoa.org/projects/ph.

The relatively large amount of archeological information compiled in the Pompeian households database supports the application of statistical data mining tools. In this paper we apply one such tool, Bayesian mixed-membership modeling, in which rooms are modeled as having mixtures of functions, and functions are modeled as distributions over a "vocabulary" of object types. Mixed-membership models have previously been applied to a broad range of inference problems in applications such as population genetics [10], text mining [3] and social network analysis [1]. We show that mixed-membership models both address several theoretical problems with simpler clustering methods and also provide improved predictive ability. Compared with a trained archaeologist, such models are naive and simplistic, but they have the advantage that they do not bring preconceived notions about culture along with them. As a demonstration, after assessing the model's predictive ability we consider several issues raised by Allison, and attempt to provide a perspective that is, if not unbiased, at least mathematically concrete in its biases.

---

[1]http://www.stoa.org/projects/ph/home

Table 1: **The probability of the top five most probable object types for eight functional groups ($K = 20$).** Groups correspond roughly to object categories such as storage furniture, decorative statuary, ceramic storage jars, and chests and cupboards.

| | | | |
|---|---|---|---|
| 0.403 | architectural door fitting | 0.359 | large sculpture/sculpture fragment |
| 0.376 | door/chest/cupboard fitting | 0.073 | fixed statue base |
| 0.031 | chest/cupboard fitting | 0.068 | fountain/fountain fitting |
| 0.023 | fixed seat | 0.059 | marble or stone basin |
| 0.020 | key | 0.056 | marble base/statue base/basin base |
| 0.165 | pottery amphora/amphoretta/hydria | 0.566 | chest/cupboard fitting |
| 0.120 | pottery amphora fragment/lid | 0.046 | glass bottle/flask/pyxis |
| 0.103 | unidentified pottery vessel | 0.043 | chest/cista |
| 0.095 | nail | 0.038 | chest fitting |
| 0.086 | pottery jar/vase | 0.028 | cupboard |
| 0.076 | measuring equipment | 0.355 | vehicle fragment |
| 0.050 | iron or lead strip | 0.309 | harness |
| 0.043 | chisel | 0.065 | pendant |
| 0.034 | pick/pickaxe | 0.029 | coin |
| 0.033 | bone fitting/strip | 0.022 | cart/wagon |
| 0.194 | glass bottle/flask/pyxis | 0.142 | pottery jug |
| 0.116 | small glass bottle | 0.090 | pottery pot |
| 0.060 | glass beaker/cup | 0.069 | ceramic lamp |
| 0.055 | small glass jar/vase | 0.048 | terra sigillata bowl/cup |
| 0.054 | jewelry | 0.039 | pottery cup/small bowl |

## 2 Clustering methods in household archaeology

There are several previous examples of data-driven methods in the archaeological study of ancient households. Fiedler considers the relative proportions of fine, cooking, and plain ceramic ware found in three room types in the Greek city of Leukos [7]. Ciolek-Torello applies PCA, factor analysis, and multidimensional scaling to objects found at Grasshopper Pueblo in Arizona [6], but finds the latent factors difficult to interpret. Cahill performs a $k$-means analysis of houses in the Greek city of Olynthus, using the percentage of floor space allocated to each room type as input variables [5].

Cahill also identifies several problems faced in attempting to recover patterns of activity from household assemblages [5]:

- Artifact types are high-dimensional. The database of houses at Olynthus contains more than 1000 distinct types, comparable to Pompeii's 240. This number of types cannot be easily studied and visualized.

- Rooms can serve more than one function, and artifact types can be used for more than one purpose. A particular type of large shallow bowl is often associated with washing, but "other examples of these artifacts, however, are found together with grinding assemblages and may have been used for kneading dough." Cahill continues "just as the use of space in ancient houses was flexible and varied according to season or need, the use of artifacts could change, and our models for interpretation must be flexible enough to allow for such changeability."

- Quantification of artifacts is critical. As an example, ancient looms used weights suspended by bundles of warp threads. These weights tend to survive after the organic matter of a loom has decayed. Cahill argues, however, that the presence of a single weight is not sufficient to indicate that a room was used for weaving, and that only the presence of a dozen or more weights in the same location indicates the presence of a loom. Both Cahill and Ciolek-Torello advocate using original data frequencies rather than binary presence/absence variables.

Bayesian mixed membership models address all three issues. First, such models are appropriate for high-dimensional categorical variables, having been applied to text documents, where vocabulary sizes are in the tens of thousands. Second, mixed membership models allow individual groups of observations to combine multiple "components" such as topics or activities, while simultaneously allowing any observed dimension, such as a word or artifact type, to be generated with non-zero probability by any component. Finally, using the multinomial event model rather than the binary event model allows us to take into account the number of times an artifact type occurs, rather than the simple fact that that type is present. For example, a single loom weight by itself might be absorbed into the functional group of other objects in the same room, while a group of 15 loom weights provides substantial evidence that a functional group with high probability of generating loom weights is active.

## 3 Models for predicting room contents

The goal of this work is to use the evidence available to us from archaeological excavations to learn predictive models for the contents of different rooms in Roman houses. Such models should then help us better understand how the Romans used their domestic spaces. The information in the database compiled by Allison consists of a set of 30 houses, each consisting of between 7 and 56 rooms. Details on the houses are provided in Table 4. Each room is labeled with a type based on its architectural features. Room types are further divided between rooms surrounding the *atrium*, an open courtyard adjacent to the main street entrance (room 01 in Figure 1), rooms surrounding a garden area (room 09), and other rooms, including kitchens and bath complexes. Descriptions of room types and their frequencies are shown in Table 2. In each room, artifacts of different types were recorded.

We now introduce mathematical notation for these entities. Let $\mathcal{H}$ be the set of houses $\{h_1, ..., h_{30}\}$. A room $r$ in house $h$ has type $t \in \{1, ..., 22\}$, and contains a set of objects. Objects are instances of a fixed dictionary of artifact classes $\mathcal{A}$. Let the set of object class indicators $x_1, ..., x_{N_r}$ be the objects in room $r$, such that $x_3 = a$ if the third object is of class $a$. For readers familiar with text processing, this representation is similar to the standard bag-of-words model, with rooms analogous to documents, and objects analogous to words. Inspired by methods that have been applied to document analysis, we present four predictive models (names used in figures are shown in parentheses).

1. A single-distribution model (Simple), equivalent to a unigram language model. This model predicts artifact types based on the frequency of previously-seen artifacts, smoothed using a Dirichlet model [12]. Let $N_a$ be the number of instances of artifact type $a$ in the training data and $\eta$ be a smoothing parameter. The probability of a room $r$ is

$$P(\boldsymbol{x}^{(r)}) = \prod_{i=1}^{N_r} \frac{N_{x_i} + \eta}{\sum_a N_a + |\mathcal{A}|\eta}. \quad (1)$$

2. A conditional type-distribution model (CondSimple), a "Naive Bayes" model. This model is identical to the previous model, but takes into account room types. Let $N_{a|t}$ be the number of instances of artifact type $a$ in rooms of type $t$ in the training data. Conditioned on its type, the probability of a room $r$ is

$$P(\boldsymbol{x}^{(r)}|t) = \prod_{i=1}^{N_r} \frac{N_{x_i|t} + \eta}{\sum_a N_{a|t} + |\mathcal{A}|\eta}. \quad (2)$$

3. A mixed-membership "topic model" over functional groups (FG). As in text-oriented topic models [3], room contents are drawn from a mixture of $k$ component distributions, which we describe here as "functional groups" (ie "topics" in document modeling). Examples are shown in Table 1. Given an allocation of objects in the training set to functional groups (either through hard or soft assignments), let $N_{a|k}$ be the number of instances of type $a$ in group $k$. The probability of an object type $a$ given a group $k$ $P(a|k, \beta)$ has the same form as Eq. 1 with $N_{a|k}$ substituted for $N_a$ and $\beta$ for $\eta$. Let $z_1, ..., z_{N_r}$ be indicator variables specifying an allocation of the objects $x_1, ..., x_{N_r}$ in a new room $r$, such that $z_i = a$ if object $x_i$ is generated by group $k$, and $N_{k|r} = \sum_i I_{z_i = k}$. Let $\alpha_1, ..., \alpha_K$ be a vector of Dirichlet parameters. The probability of a room is then a marginalization over all possible settings of $\boldsymbol{z}$:

$$P(\boldsymbol{x}^{(r)}) = \sum_{\boldsymbol{z}} \frac{\Gamma(\sum_k \alpha_k)}{\Gamma(\sum_k \alpha_k + N_r)} \prod_k \frac{\Gamma(\alpha_k + N_{k|r})}{\Gamma(\alpha_k)}$$
$$\times \prod_{i=1}^{N_r} P(x_i | z_i, \beta). \quad (3)$$

4. A mixed-membership model over functional groups conditioned on room type (CFG). This model is similar to the previous model, but with a distinct vector of Dirichlet parameters $\alpha_1^{(t)}, ..., \alpha_K^{(t)}$ for each room type $t$.

These first two models draw the entire contents of a room from a single distribution. The second two are mixed-membership models that draw objects in a room from a room-specific mixture of distributions. The first and third models do not take into account room types, while the second and fourth use room types at training time and condition on room type at testing time.

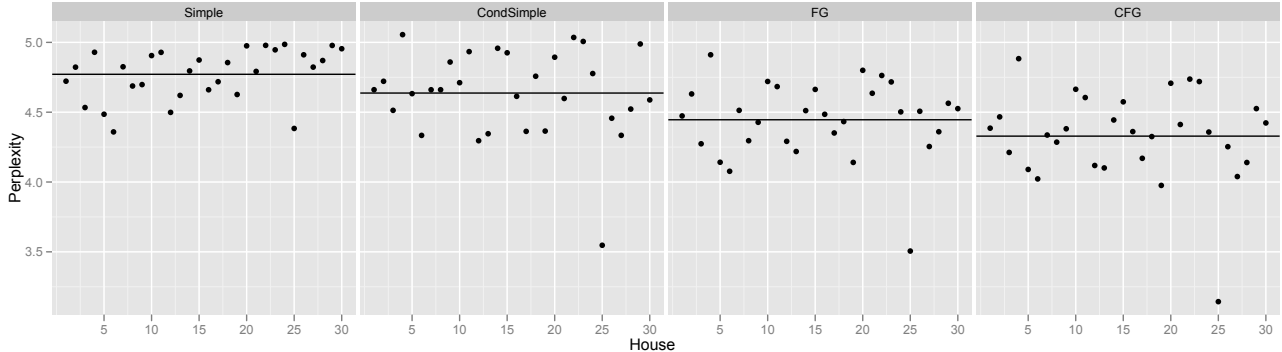

Figure 2: **The mixed-membership model conditioned on room types has the best perplexity.** Each point represents the average perplexity over all rooms of each of the 30 houses. The dark line shows the mean average perplexity for each model.

Inference in the first two models can be accomplished by counting. Inference in the mixed-membership models was carried out by Gibbs sampling using the Mallet toolkit [8]. For the simple mixed-membership model (FG), asymmetric room-group hyperparameters $\alpha_1, ..., \alpha_k$ and a single symmetric hyperparameter $\beta$ were iteratively optimized as advocated by Wallach et al. [11]. Room-type-specific Dirichlet priors over functional groups were learned post-hoc from the simple mixed-membership model given saved Gibbs sampling states [9].

## 4 Evaluation of Predictive Ability

We begin by evaluating these models using leave-one-out cross validation on houses. For thirty training runs, we estimated parameters for the four models. For the mixed-membership model, we trained 25 models, five random initializations each for $K \in \{10, 15, 20, 25, 30\}$. We ran each model for 10,000 iterations of Gibbs sampling, saving states every 500 iterations.

Eq. 3 involves an intractable summation over an exponential number of possible functional group assignments. Held-out probability for rooms in the left-out houses was therefore estimated using Buntine's sequential method [4]. In order to make comparisons between rooms with differing numbers of objects, we report perplexity, the negative log probability of the room divided by the number of objects, as is typical in language modeling.

The choice of the number of functional groups $K$ affected perplexity, with more groups leading to greater likelihood. Average perplexities for different values of $K$ are shown in Table 3. We did not consider $K$ greater than 30 due to the relatively small size of the data set. All subsequent perplexity numbers are averaged over these five values of $K$.

Table 3: **CFG has better perplexity than FG.** Perplexity improves with larger $K$.

| Model | 10 | 15 | 20 | 25 | 30 |
|---|---|---|---|---|---|
| FG | 4.49 | 4.48 | 4.46 | 4.46 | 4.45 |
| CFG | 4.41 | 4.38 | 4.35 | 4.33 | 4.32 |

Overall, the conditional functional groups mixed-membership model provides the best predictive performance averaged over all rooms, followed by the functional groups model, the conditional type-distribution model, and the simple single-distribution model. The difference in means between each pair of models is significant according to a pairwise $t$-test. We found small $p$-values for all pairs except between the two non-mixed-membership models, which have $p = 0.002$. This performance of different models, however, varies between houses and between room types.

### 4.1 Perplexity by house

Figure 2 shows the average perplexity of each house under the four models. In most cases the mixed-membership model conditioned on room type performs best, followed by the simple mixed-membership model. The conditional single-distribution model outperforms the simple "unigram" model in the majority of cases, 21 of 30 houses, but not always. House 4, the Casa dei Ceii, is the most unpredictable in all models. Allison reports that this house shows signs of having been disturbed after the eruption, possibly by people familiar with the floor plan of the house and the neighboring house. The most well-predicted house is number 25, House VIII 2,26, which has the fewest objects of any house (17). There is no clear relationship between average perplexity and number of objects, however: the largest collection (house 9, the Casa del Menandro), is in the middle range of perplexity scores.

Table 2: **Room types specified by Allison [2]**. These differ slightly from traditional nomenclature using Latin names derived from literary sources.

| ID | N | Sect. | Description | Latin name |
|----|----|-------|-------------|------------|
| 1 | 18 | front | main entranceway | *fauces, vestibula* |
| 2 | 8 | front | room leading directly off from entranceway | *cella ostiaria* |
| 3 | 35 | front | front hall, usually with central opening and pool | *atrium* |
| 4 | 80 | front | small closed room off side of front hall | *cubiculum* |
| 5 | 16 | front | open-fronted area off side of front hall | *ala* |
| 6 | 24 | front | large/medium room off corner of front hall | *triclinium* |
| 7 | 18 | front | open-sided room opposite main entrance or leading to garden | *tablinum* |
| 8 | 25 | front | long, narrow internal corridor | *fauces, andrones* |
| 9 | 36 | garden | main garden, collonaded garden | *peristylum,* etc. |
| 10 | 17 | garden | large/medium closed room off garden/terrace with no view | *triclinium* |
| 11 | 31 | garden | large/medium open-fronted room off garden/terrace with window or wide entranceway | *oecus, exedra, triclinium* |
| 12 | 43 | garden | small closed room off garden/terrace or lower floor | *cubiculum* |
| 13 | 9 | garden | small open-fronted area of garden/terrace or lower floor | *exedra* |
| 14 | 42 | other | room with cooking hearth (kitchen) | *culina* |
| 15 | 11 | other | latrine as entire room | *latrina* |
| 16 | 43 | other | other room outside main front-hall/garden complex | *repositorium,* etc. |
| 17 | 23 | other | stairway | |
| 18 | 14 | other | secondary internal garden or court, usually not collonaded | *hortus* |
| 19 | 4 | other | secondary entrance or entrance courtyard | *fauces* |
| 20 | 7 | other | room at front of house open to street (shop) | *tabernae* |
| 21 | 11 | other | bath area | *balneae* etc. |
| 22 | 53 | other | upper floor rooms and material in upper-level deposits | *cenaculum* |

### 4.2 Perplexity by room type

We can also look at the distribution of perplexities by room type. There are clear and consistent architectural patterns in Roman houses, but architectural similarity does not necessarily imply similarity of use. Evidence from Pompeii is commonly used to infer how Romans organized their household activities. It is therefore critical to evaluate how consistent the contents of rooms are between each other, in order to determine whether we can extrapolate from the sample of available rooms to more general conclusions about the specific function of rooms.

Allison's classification of room types is shown in Table 2. Average perplexities over each room type are shown in Figure 3 (note that there is an additional type 0 for lower floors included in the online database but not the published description). The dark line shows the mean of the average perplexities for room types. In most cases, the two mixed-membership models outperform the non-mixed-membership baseline models, indicating that rooms are better described by a combination of more specific distributions learned from several different room types than room-specific distributions. The relative performance of the four models, however, varies between room types. Certain specialized rooms, such as entranceways (1), kitchens (14), latrines (15) and stairways (20) are better predicted by the two conditional models than their non-conditional counterparts. The type-distribution model is better than the functional group model for kitchens ($p = 0.0002$), but not significantly for front halls (3, $p = 0.62$). In many other cases, however, the conditional type-distribution model is either indistinguishable from or worse than the single-distribution model.

If the average perplexity for a room type is greater for a non-conditional model than that model's conditional version, room type information is useful for prediction, implying that the contents of a particular room type are similar across houses. If perplexity is greater for the conditional model, individual instances of a room type do not predict the contents of similar rooms. The variability of perplexities are generally smaller for rooms centered around the front hall and the garden than "other" specialized rooms. For example, kitchens (14) and stairways (17) are better predicted conditioned on their types, indicating that these rooms have consistent contents. This argument is to some extent circular, because these two room types are at least partly defined by the presence of large, immovable fixtures such as hearths and stairs, but it provides an indication of what the model should show if there is consistent use. In contrast, rooms open to the street (20) are *less* well predicted when conditioned on their type, indicating that conditional models are overfitting, and suggesting that we should be wary of generalizing about the contents of such rooms given our sample.

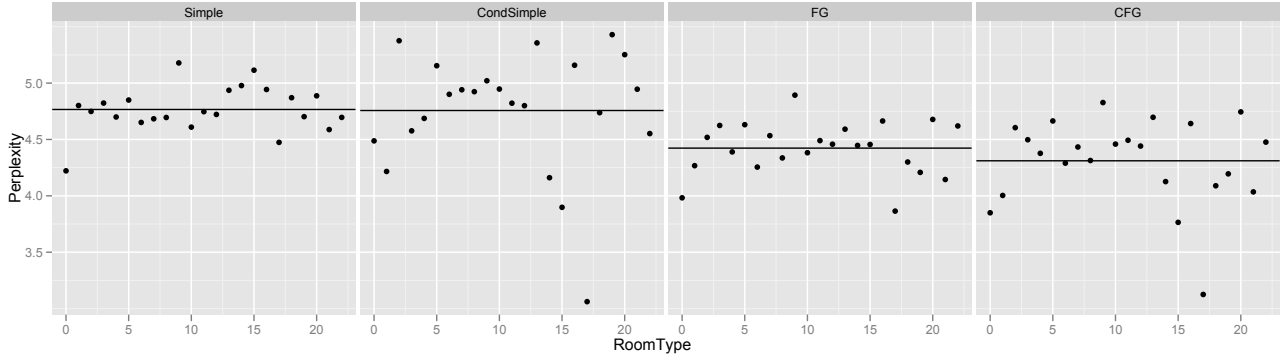

Figure 3: **Conditioning on room type frequently does not improve performance.** Each point shows the average perplexity for one of the 22 room types (see Table 2). Main gardens (9) were difficult for all models to predict, indicating varied usage, while specialized rooms such as kitchens (14), latrines (15), and stairways (17) are distinctive and similar across houses. The dark line shows the mean average perplexity for each model.

## 5 Context-driven estimation of artifact function

The models presented in Section 3 provide a tool for testing archaeological questions. In this section we attempt to validate two hypotheses proposed by Allison.

When artifacts are excavated, standard archeological practice involves removing them to secure storage for preservation. Although the location of artifacts is carefully noted in modern scientific digs, artifacts in storage tend to be analyzed in comparison to typologically similar objects rather than within their original context. As a result, questions about the use or function of particular artifact types have been based more on arbitrary tradition and researchers' perceptions of what an artifact is reminiscent of than the types of objects found with the object. For example, Allison identifies two classes of artifact, the *casseruola* ("casserole dish") and *forma di pasticceria* ("pastry mold"), that were named based on similarities to 19th century household objects. She cites research going back more than 100 years suggesting that these items were not used in food preparation contexts, but finds that contemporary scholars still make this assumption due to their (modern) names.

Using imaginative 19th century names to understand the function of objects is not good practice, but simply counting cooccurrences within rooms may also lead to improper rejection of connections. Due to small sample sizes, it may be possible for two related items to never appear together in a room purely by chance. In order to reduce bias of both kinds, we explore the function of these artifact types using only cooccurrence data, without any reference to the actual typology of the objects. We use the functional group mixed-membership model to detect clusters of object cooccurrence that may indicate functions. Note that we are still dependent on experts to classify physical objects into appropriate categories, but given those classifications we make no further archeological assumptions in training the model. At the same time, we expect that if two artifact types $a_1$ and $a_2$ appear with artifacts $a_3, a_4$ and $a_5$, all five objects will tend to be placed in the same functional group, even if $a_1$ and $a_2$ never occur together.

Intuitively, if two objects share a similar pattern of use, they should both have high probability in one or more "topic". Given a model, we can evaluate the probability that two artifact types $a_1$ and $a_2$ will be produced by the same functional group, marginalized over functions, as $P(a_1, a_2) = \sum_k P(k) P(a_1 \mid k) P(a_2 \mid k)$, where $P(k)$ is proportional to the average number of tokens assigned to functional group $k$ and $P(a_1 \mid k)$ is proportional to the average number of objects of type $a_1$ assigned to functional group $k$. Table 6 shows results for the two types mentioned previously. There is little to no connection to food preparation objects, supporting Allison's claim that modern names for these items are incorrect.

Table 4: **The 30 houses in the data set vary in the number of rooms $R$ and number of of objects recorded $N$.** Houses in region VIII were excavated earliest, when records were not kept as rigorously as in modern excavations.

| ID | $R$ | $N$ | Location | Name |
|---|---|---|---|---|
| 1 | 15 | 189 | I 6,4 | Casa del Sacello Iliaco |
| 2 | 7 | 63 | I 6,8-9 | House I 6,8-9 |
| 3 | 12 | 275 | I 6,11 | Casa dei Quadretti Teatrali |
| 4 | 12 | 125 | I 6,15 | Casa dei Ceii |
| 5 | 16 | 67 | I 6,13 | Casa di Stallius Eros |
| 6 | 13 | 99 | I 7,7 | Casa del Sacerdos Amandus |
| 7 | 25 | 353 | I 7,10-12 | Casa dell'Efebo |
| 8 | 20 | 112 | I 7,19 | House I 7,19 |
| 9 | 56 | 886 | I 10,4 | Casa del Menandro |
| 10 | 15 | 522 | I 10,7 | Casa del Fabbro |
| 11 | 16 | 164 | I 10,8 | House I 10,8 |
| 12 | 16 | 148 | I 10,11 | Casa degli Amanti |
| 13 | 17 | 178 | I 11,6 | Casa della Venere in Bikini |
| 14 | 18 | 122 | III 2,1 | Casa di Trebius Valens |
| 15 | 52 | 748 | IX 13,1-3 | Casa di Julius Polybius |
| 16 | 30 | 242 | V 2,i | Casa delle Nozze d'Argento |
| 17 | 17 | 145 | V 4,a | Casa di M. Lucretius Fronto |
| 18 | 16 | 176 | VI 15,1 | Casa dei Vettii |
| 19 | 17 | 158 | VI 15,5 | House VI 15,5 |
| 20 | 13 | 158 | VI 15,8 | Casa del Principe di Napoli |
| 21 | 20 | 187 | VI 16,7 | Casa degli Amorini Dorati |
| 22 | 10 | 141 | VI 16,15 | Casa della Ara Massima |
| 23 | 19 | 144 | VI 16,26 | House VI 16,26 |
| 24 | 23 | 120 | VIII 2,14-16 | House VIII 2,14-16 |
| 25 | 10 | 17 | VIII 2,26 | House VIII 2,26 |
| 26 | 11 | 61 | VIII 2,28 | House VIII 2,28 |
| 27 | 20 | 61 | VIII 2,29-30 | House VIII 2,29-30 |
| 28 | 20 | 71 | VIII 2,34 | House VIII 2,34 |
| 29 | 31 | 168 | VIII,2,39 | Casa di Giuseppe II |
| 30 | 17 | 63 | VIII 5,9 | House VIII 5,9 |

Table 5: Objects most likely to occur with *bronze casseruola*, marginalized over 20 functional groups ("topics"): **the first item associated with food preparation (found in kitchens with evidence of exposure to fire) is well down the list.**

| | **bronze casseruola** |
|---|---|
| 0.00036 | door/chest/cupboard fitting |
| 0.00035 | glass bottle/flask/pyxis |
| 0.00028 | small glass bottle |
| 0.00026 | pottery jug |
| 0.00024 | bronze jug/jug fragment |
| 0.00022 | chest/cupboard fitting |
| 0.00020 | ceramic lamp |
| 0.00018 | jewelry |
| 0.00016 | pottery beaker/small vase |
| 0.00015 | coin |
| 0.00013 | pottery pot |
| | ... |
| 0.00010 | table/table fittings/table base |
| 0.00010 | pottery jar/vase |
| 0.00010 | **bronze cooking pot/basin/pot/fragment** |

Table 6: Objects likely to occur with a *forma di pasticceria* are shown on the right, again showing **no significant connection to food preparation**.

| | **silver vessel/forma di pasticceria** |
|---|---|
| 0.00008 | jewelry |
| 0.00004 | silver cup/bowl/cup fragment |
| 0.00004 | chest/cupboard fitting |
| 0.00004 | silver patera/casseruola/plate |
| 0.00003 | casket fitting |
| 0.00002 | coin |
| 0.00002 | bronze jug/jug fragment |
| 0.00002 | door/chest/cupboard fitting |
| 0.00002 | bronze or silver spoon |
| 0.00002 | chest fitting |
| 0.00001 | box/casket |
| 0.00001 | part of coin hoard |
| 0.00001 | small glass bottle |
| 0.00001 | hair pin |
| 0.00001 | ceramic lamp |

## 6 Modeling room functions

The Roman houses included in the database show strong architectural patterns. Much of the study of Pompeian households has involved identifying categories of rooms and assigning functions to them. Again, Allison argues that commonly held assumptions about such functions are incorrect. For example, the *atrium* is often described as a formal space, in which the *pater familias* received his clients and distributed gifts from a large metal chest. Allison argues that the *atrium* was more of a utilitarian, industrial space. Similarly, a closed room off the *atrium* is described as a *cubiculum* ("bedroom"), but there is little evidence that these rooms correspond to modern concepts of bedrooms.

Before using an analysis of objects to provide information about the function of spaces, it is important to establish whether architectural room types have consistent patterns of object contents. The city suffered a severe earthquake 17 years before the final eruption, and was disrupted before and after the eruption. As shown previously, certain types of rooms have strong connections to particular artifact classes, while for others, conditioning on room type does not improve predictive performance.

As a result of this variability of room contents, it is difficult to make general claims about what activities occurred in which spaces. We can, however, attempt to rule out certain possibilities. Table 7 shows probable functional groups for several architectural types. *Atria* show evidence of utilitarian storage: large ceramic amphorae. The only type that shows significant evidence of objects related to bedding is type 11 (large open rooms with views of the garden); the *cubiculum* does not.

Table 7: **The *cubiculum* is not a bedroom**. Distributions of functional groups in common room types.

|       | Type 3 (*atrium*) |
|-------|-------------------|
| 0.279 | pottery amphora/amphoretta/hydria, building material, stairway, impluvium/compluvium, puteal/puteal fragment |
| 0.108 | chest/cupboard fitting, chest/cista, glass bottle/flask/pyxis, chest fitting, cupboard |
| 0.083 | coin, ceramic lamp, small glass bottle, architectural fitting, jewelry |
| 0.077 | pottery jug, ceramic lamp, table/table fittings/table base, terra sigillata bowl/cup, seashell/conch/snail shell |
|       | **Type 4 (*cubiculum*)** |
| 0.122 | recess, built-in cupboard, stairway, niche, unidentified fixture/mound |
| 0.116 | shelving/mezzanine/suspension nails, recess, pottery amphora/amphoretta/hydria, cistern head, pottery pot |
| 0.092 | coin, ceramic lamp, small glass bottle, architectural fitting, jewelry |
| 0.065 | pottery jug, pottery amphora/amphoretta/hydria, pottery pot, pottery jar/vase, pottery plate/dish/tray |
|       | **Type 11 (*exedra, oecus, triclinium*)** |
| 0.111 | door/chest/cupboard fitting, coin, ring, chest/cupboard fitting, nail |
| 0.109 | recess, built-in cupboard, stairway, niche, unidentified fixture/mound |
| 0.080 | bed/couch fragment, door/chest/cupboard fitting, large sculpture/sculpture fragment, architectural door fitting, **bed/couch** |
| 0.058 | human skeleton, jewelry, part of coin hoard, bed/couch, bag |

## 7 Conclusions

The excavations at Pompeii provide a unique view into domestic life in 1st century Roman Italy. We analyzed a database of 30 Roman houses using four predictive models, comprising all combinations of single-distribution and mixed-membership models, and unconditional models and models that condition on Allison's room type annotations. Building statistical models of the distribution of objects in these houses is valuable in that it provides a data-driven method for measuring how confident we can be in making claims about the function of domestic spaces.

In many cases, conditioning on room type did not improve predictive performance, demonstrating that room function is not necessarily determined by architectural characteristics. Our results are consistent with Allison's challenge to assumptions that small closed rooms (*cubicula*) were used for sleeping and that two object types (*casseruola* and *forma di pasticceria*) were used in food preparation.

We find that mixed-membership models provide a promising method for identifying patterns of use in archaeological data. They find interpretable functional groups, have good predictive performance, and address theoretical problems with simpler clustering methods. As the quantity of digitized data from excavations increases, such methods should provide a powerful tool for interpreting archaeological data.

**Acknowledgements**

The author is supported by a Google Digital Humanities Research grant. Eric Poehler suggested the data set. Ross Scaife contributed to the development of the Pompeian Households database. We hope he would have liked this paper.